\documentclass[letterpaper]{article} 
\usepackage{aaai23}  
\usepackage{times}  
\usepackage{helvet}  
\usepackage{courier}  
\usepackage[hyphens]{url}  
\usepackage{graphicx} 
\usepackage{natbib}  
\usepackage{caption} 
\usepackage{algorithm}
\usepackage{algorithmic}

\usepackage{tikz}
\usepackage{float}
\usepackage{multirow}
\usepackage{comment}
\usepackage{amsmath,amssymb} 
\usepackage{color}
\usepackage{diagbox}
\usepackage{makecell}

\usepackage{newfloat}
\usepackage{listings}
\DeclareCaptionStyle{ruled}{labelfont=normalfont,labelsep=colon,strut=off} 
\floatstyle{ruled}
\newfloat{listing}{tb}{lst}{}
\floatname{listing}{Listing}
\title{CL3D: Unsupervised Domain Adaptation for Cross-LiDAR 3D Detection}
\author{
    Xidong Peng, \textsuperscript{\rm 1}
    Xinge Zhu, \textsuperscript{\rm 3}
    Yuexin Ma \textsuperscript{\rm 1,2 \footnote{Corresponding author}}
}
\affiliations{
    \textsuperscript{\rm 1}ShanghaiTech University\\
    \textsuperscript{\rm 2}Shanghai Engineering Research Center of Intelligent Vision and Imaging
    \textsuperscript{\rm 3}The Chinese University of Hong Kong\\

    \{pengxd, mayuexin\}@shanghaitech.edu.cn
}

\usepackage{bibentry}

\begin{document}

\maketitle

\begin{abstract}
Domain adaptation for Cross-LiDAR 3D detection is challenging due to the large gap on the raw data representation with disparate point densities and point arrangements. By exploring domain-invariant 3D geometric characteristics and motion patterns, we present an unsupervised domain adaptation method that overcomes above difficulties. First, we propose the Spatial Geometry Alignment module to extract similar 3D shape geometric features of the same object class to align two domains, while eliminating the effect of distinct point distributions. Second, we present Temporal Motion Alignment module to utilize motion features in sequential frames of data to match two domains. Prototypes generated from two modules are incorporated into the pseudo-label reweighting procedure and contribute to our effective self-training framework for the target domain. Extensive experiments show that our method achieves state-of-the-art performance on cross-device datasets, especially for the datasets with large gaps captured by mechanical scanning LiDARs and solid-state LiDARs in various scenes. Project homepage is at
\url{https://github.com/4DVLab/CL3D.git}.
\end{abstract}

\section{Introduction}

 \begin{figure}[t]

    \centering
    \includegraphics[scale=0.45]{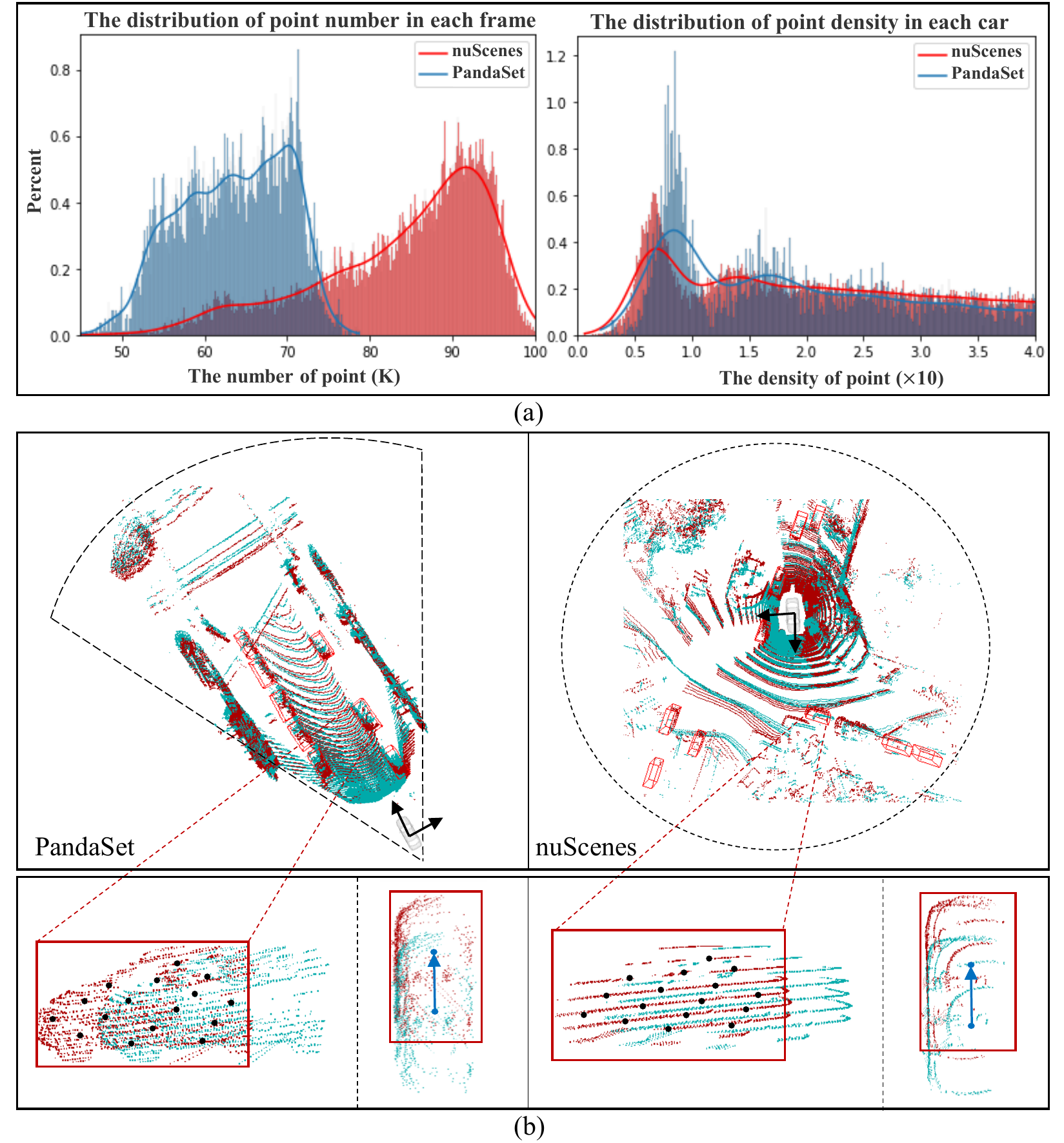}
  \caption{We show several key domain gaps between mechanical scanning LiDAR (from nuScenes) and solid-state LiDAR (from PandaSet). (a) shows the comparison of statistics of point numbers and point densities. (b) describes domain gaps of perception range and instance-level point distribution. Two adjacent frames are shown in green and red. The geometric structure represented by the black sampling points in the side-view and the motion pattern in the bird-eye-view are domain-invariant for cars.}
    \label{fig_teaser}
\vspace{-2ex}
\end{figure} 

Due to the advantages of accurately capturing depth information of large-scale scenes, LiDARs become crucial sensors for the 3D perception of autonomous driving and robotics. Boosted by deep learning techniques, LiDAR-based 3D detection~\cite{yan2018second,zhu2021cylindrical,qi2019deep,zhu2020ssn,chen2020hierarchical,shi2020point,zhang2022not,Cong_2022_CVPR,hou2022point} has made great progress and becomes the main solution for many autonomous driving companies. However, deep learning-based methods rely heavily on massive annotated data, which is time-consuming and expensive, especially for labeling 3D point clouds of large scenes. Moreover, the domain adaptation for point clouds gets more challenging compared to image-based adaptation tasks~\cite{chen2018domain,saito2019strong,zhu2018penalizing,cui2020gradually,li2022stepwise} where the image-based domain gap is more about the explicit appearance including lighting and weather while the domain gap existing in LiDAR is mainly reflected in the raw point representation, which causes image-based domain adaptation strategies inapplicable. 

Different types of LiDAR have obvious difference on the point representation. For the widely-used mechanical scanning LiDARs, they have various beams with different point densities. Compared with them, solid-state LiDARs are based on another principle of physics, having disparate perception ranges, point densities, and point arrangements. We show the statistics and visualization to describe the domain gaps in Figure.~\ref{fig_teaser}. Actually, solid-state LiDARs become more and more popular because it is cheaper and has longer perception distance and longer longevity, which makes autonomous vehicles installed with them more feasible for quantity production. However, most current open large-scale 3D datasets~\cite{geiger2012we,caesar2020nuscenes,sun2020scalability} are captured by mechanical LiDARs. How to solve the domain adaptation problem between these two distinct LiDARs becomes extremely urgent and significant.  Previous works~\cite{yang2021st3d,wang2020train,luo2021unsupervised,zhang2021srdan} mainly consider the consistent objects' sizes for the domain alignment. Actually, the more important domain-invariant features in 3D point cloud are geometric characteristics including shapes, scales, etc., and temporal motion information. For example, sedan cars bear resemblance in shape and motion representation; the poses and actions of pedestrians are also alike. 

Motivated by this, we propose \textbf{CL3D}, an unsupervised domain adaptation method for LiDAR-based 3D detection, especially for solving the cross-device setting, via exploring the spatial geometry information and temporal motion representation. Specifically, we develop a framework consisting of two key components, \textbf{Spatial Geometry Alignment (SGA)} and \textbf{Temporal Motion Alignment (TMA)}, to extract the geometric features of both local structure and global context, and model the motion pattern in consecutive point clouds, respectively. Based on SGA and TMA, a prototype representation with geometry and motion constraints for each specific class is proposed, where the similarity between current sample and average target prototype (updated via exponential moving average) is used to reweight the confidence of pseudo labels generated by the self-training framework. Different from existing rigid pseudo-label selection process, this proposed soft-selection mechanism could reduce the effect of incorrect labels and avoid directly discarding correct labels with low confidence. In addition, we explore several strategies to solve the perception range gaps between mechanical scanning LiDARs and solid-state LiDARs and give conclusions to facilitate the data pre-processing.

We conduct extensive experiments on various cross-LiDAR and synthetic-to-real domain adaptation tasks, and all get state-of-the-art performance. We also conduct detailed ablation studies quantitatively and qualitatively to demonstrate the effectiveness of different modules of our method. To our knowledge, we are the first to explore the point cloud-based 3D detection domain adaptation for the challenging mechanical-to-solid-state cross-LiDAR setting.

Our contributions are summarized as follows.

\begin{itemize}
    \item We investigate the cross-LiDAR domain gap and propose an unsupervised domain adaptation method for 3D detection, especially for the challenging mechanical to solid-state LiDAR adaptation.
    
    \item We propose SGA and TMA to extract similar 3D shape geometric characteristics and motion patterns belonging to objects of the same category to align two domains by reweighting pseudo labels with soft constraints.
    
    \item Our method achieves state-of-the-art performance on unsupervised domain adaptation for cross-LiDAR 3D Detection.
\end{itemize}

\section{Related Work}
\noindent\textbf{LiDAR-based 3D Object Detection}\quad Due to the advantages of capturing depth information in large-scale scenes, more and more LiDAR-based methods for 3D object detection are proposed in recent years. These methods can be divided into point-based methods and grid-based methods. The former \cite{shi2019pointrcnn,yang20203dssd,qi2019deep,chen2020hierarchical,shi2020point,he2022svga} directly extracts the features from unordered raw point clouds data with PointNet~\cite{qi2017pointnet} or PointNet++~\cite{qi2017pointnet++} to generate 3D proposals. These methods preserve the original geometric features of point cloud but become time-consuming for processing large-scale outdoor data. The latter \cite{he2020structure,shi2020points,shi2020pv,yan2018second,zhu2021cylindrical,zhu2020ssn,hu2022afdetv2,zhang2022not} utilizes structured representation to quantize the LiDAR data into the fix-sized voxel or pillar grids and then use this representation to further extract semantic features for 3D object detection. Such methods perform well in efficiency and are usually adopted in autonomous driving scenarios.  In our work, we chose the state-of-the-art grid-based 3D detector CenterPoint~\cite{yin2021center} as base network, which is a one-stage anchor-free method with high efficiency and accuracy. Instead of generating a new detector, we aim at adding new modules to basic detector to adapt it to the unsupervised domain adaptation task.

\noindent\textbf{Domain Adaptation for 2D image}\quad There are already a lot of investigations \cite{hoffman2018cycada,chen2018domain,saito2019strong,zhu2018penalizing,li2022stepwise,yu2022sc} about domain adaptation for various 2D computer vision tasks such as image-based detection and segmentation. Many domain adaptation methods \cite{hoffman2016fcns,ganin2016domain,bousmalis2016domain,cui2020gradually,hu2020unsupervised,yang2020mind} use adversarial learning to align feature distributions across different domains inspired by GANs \cite{goodfellow2014generative}, while some statistic-based methods \cite{mancini2018boosting,maria2017autodial,long2017deep,sun2016deep,xu2020reliable,long2015learning} employ the statistic-based metrics to the domain gap between two different data distributions. Moreover, pseudo-label-based self-training is becoming a more and more popular approach~\cite{khodabandeh2019robust,roychowdhury2019automatic,seibold2022reference,yao2022enhancing} for unsupervised domain adaptation, which is easier to implement compared with previous two kinds of methods. Our method also adopts such self-training mechanism for the target domain. However, unlike images with unchanged regular pixel representations for different environments and devices, LiDAR point cloud is changing apparently on the raw data representation with diverse point densities and distributions, which is not applicable to directly extend the image-based adaptation methods to LiDAR point cloud. Our method fully explores the specific properties of 3D data and achieves superior performance.

\noindent\textbf{Domain Adaptation for 3D point cloud}\quad To bridge the domain gap on point clouds captured in various environments and by different LiDAR sensors, some works appear recently for shape classification~\cite{qin2019pointdan}, semantic segmentation~\cite{wu2018squeezeseg,yi2021complete,jaritz2020xmuda,xiao2022transfer,he2021re} and 3D detection~\cite{hegde2021attentive,luo2021unsupervised,xu2021spg,you2022exploiting}. We focus on the 3D detection task. To align geometry features, \cite{yihan2021learning} transforms the 3D feature to the bird's eye view with regular structures as images. However, dimensionality reduction usually misses detailed 3D geometry characteristics and is not friendly for small objects. \cite{wang2020train} utilizes the object size statistics of two domains to narrow the gap, but the performance relies on the source and target data distributions. SRDN~\cite{zhang2021srdan} aligns the features according to the instance sizes and distances to the LiDAR sensor. It is reasonable for the data captured by the same LiDAR, but not applicable for the cross-device data, where even the same object at the same location will be presented differently with different densities and arrangements. Compared with these works using the object size, our work considers a more important geometric characteristic, the shape, consisting of both the size and structure information, as well as the temporal features to pull two domains closer. ST3D~\cite{yang2021st3d} generates pseudo labels of the target domain by the model trained on source domain and then selects high-quality pseudo labels for self-supervision. However, the threshold-based method for pseudo-label selection usually leads to involving incorrect labels that have high confidence and discarding correct labels with low confidence in the self-training procedure, which may mislead the network. Our method uses a soft-selection mechanism for pseudo labels by reweighting the confidence, which can alleviate above situations to a certain extent.

\section{Methodology}
\begin{figure*}[h]
    \centering
    \includegraphics[scale=0.5]{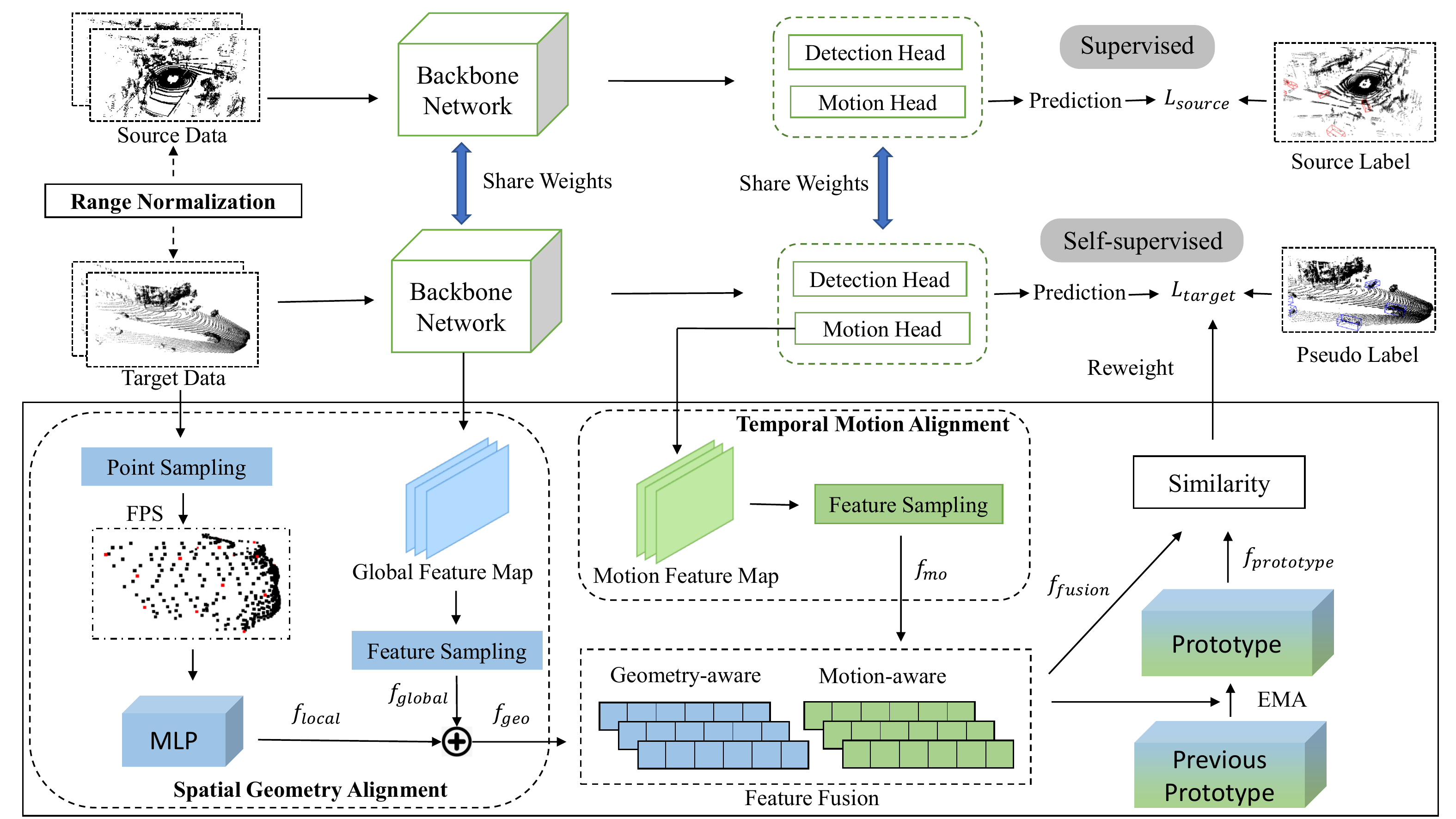}
    \caption{The pipeline of our method CL3D. Firstly, a 3D detection network is trained on source data to generate the pseudo label on target domain. During re-training on target domain, Spatial Geometry Alignment and Temporal Motion Alignment are used to extract the domain-invariant geometry-aware and motion-aware features, which are used to generate prototype for the soft selection of pseudo labels. Moreover, Range Normalization is designed to ease the range gap for better pre-processing.}
    \label{fig_framework}
\end{figure*}

\subsection{Problem Statement}

Given the source domain, denoted as $D_S = \{P^S, L^S\}$ with point cloud data $P^S$ and labeled samples $L^S$, and the target domain $D_T = \{P^T\}$ without any annotation, our task is to train a 3D detector that can generalize well to the target domain, utilizing the data in both domains. Specifically, $P^S$ and $P^T$ are captured by different types of LiDAR in various environments with different numbers, densities, arrangements, and ranges of points. 

\subsection{Framework Overview}

To align two distinct point cloud domains, we fully explore the domain-invariant features existing in raw data representations, including the geometric characteristics and motion patterns. We design two important modules in CL3D. The first is \textbf{Spatial Geometry Alignment (SGA)}, which extracts geometric features of both local structure and global context for objects. The second is \textbf{Temporal Motion Alignment (TMA)}, which aims at learning the motion features of the same kind of objects from consecutive data. Then, we generate a \textbf{feature-fusion prototype} representation based on SGA and TMA to align the source domain and target domain with geometry and motion constraints. Specifically, the target prototype is updated via exponential moving average (EMA) and the similarity between current sample and prototype is used to reweight the confidence of pseudo labels to optimize the detection results via this soft-selection mechanism. In particular, the perception range also differs a lot for mechanical scanning LiDAR and solid-state LiDAR. We further propose an effective strategy of \textbf{range normalization} for better pre-processing. The pipeline of our method, CL3D, is illustrated in Figure \ref{fig_framework}. 

\subsection{Spatial Geometry Alignment}

Different from the dense and regular representation of image pixels, points captured by LiDARs appear sparsity-varying and unordered distributions in 3D space. For various kinds of LiDAR, like widely-used mechanical scanning LiDARs with 32, 63, and 128 beams and solid-state LiDARs, there exist obvious differences in the raw data representations, leading to huge domain gaps. However, the object's geometric structure is invariant even in diverse environments and captured by disparate devices, which inspires us to extract the essential geometry features from arbitrary point cloud representations. By the 3D detection backbone trained on the source domain, we can obtain pseudo labels to localize potential objects of the target domain. To eliminate the influence brought by diverse densities and arrangements of points and make the shape feature more robust, we adopt the farthest point sampling (FPS) method to uniformly collect 16 points from the point cloud belonging to each object as the shape representation. After that, we normalize these points by transforming the coordinates from the LiDAR coordinate system to the self-coordinate system. A two-layer multi-layer perceptron (MLP) is attached to extract the point-wise geometry feature, $f_{local}$, for the local structure of objects. Considering that feature map generated by the detection backbone involves the features of the global context, which contains high-level geometry information, we sample on the feature map by interpolation to get the $f_{global}$. Then we obtain the final geometric feature $f_{geo}$ for each object by $f_{geo}=\text{concat}\{f_{local}, f_{global}\}$, which contains both local structure and global context information.

\subsection{Temporal Motion Alignment}

Additionally, the dynamic properties in objects' movement are also invariant across various domains, which can benefit the domain alignment as well. First, we input two consecutive frames of LiDAR point cloud to acquire dynamic information. Then, we add an extra motion head for the backbone network to extract the temporal feature and get the motion feature map. Finally, we conduct the feature sampling on the motion map according to the pseudo label to obtain the corresponding motion feature $f_{mo}$ of each object. Both of the features obtained by SGA and TMA will be used to generate prototype to align two domains by prototype learning.

\subsection{Prototype Learning}

In previous works, rigid filtering methods \cite{khodabandeh2019robust,roychowdhury2019automatic,yang2021st3d} are popular for the pseudo-label-based self-supervised frameworks. Commonly, rigid filtering methods are used by setting different confidence thresholds to filter prediction results, but it may mistakenly exclude some correct but low-confidence predictions. Therefore, we hope to calculate and learn the corresponding prototype representation for each specific class in the re-training process, then reweight the information from each pseudo-label by comparing the corresponding feature of pseudo-label with the prototype representation, in order to achieve a soft limitation effect for incorrect pseudo labels in the re-training process.

Usually, the quality of pseudo label on target data would dominate the effectiveness of the network on the target domain in the self-supervised framework. Particularly, although there exist large variances between different types of LiDARs, these instance-level geometry structure and temporal motion information are consistent. To this end, we employ the spatial geometry alignment and temporal motion alignment acting as a soft selection mechanism for the pseudo labels as shown in the half bottom of Figure \ref{fig_framework}. In this way, only pseudo labels with high similarity between prototypes (current labels and template prototypes) are remained.

\noindent\textbf{Feature Fusion for Prototype.} \quad In the traditional 2D object detection \cite{jiang2018learning,yang2018robust}, it is considered that the extracted semantic features of specific class from the backbone network are similar, which can be used to compute the prototype representation. However, for LiDAR-based 3D target detection, the input data is point cloud, where no texture and color information exists but the shape-aware geometric representation and dynamic motion information are presented. Therefore, we process and obtain the underlying consistent shape information and temporal motion pattern for each object to match different domains. 

Based on the geometry feature $f_{geo}$ acquired from SGA and the motion feature $f_{mo}$ obtained from TMA, we can update the prototype representation in each training iteration. Instead of averaging the fused feature information directly to obtain the prototype, we adjust the weights of these fused features by the confidence $d$ corresponding to each pseudo label as $f_{fusion} = \frac{1}{N} \sum_{i = 1}^N d_i \cdot \text{concat} \{f_{geo}^i, f_{mo}^i\}$, where N means the number of sampled features. Additionally, the prototype computed in an iteration is combined with the previous prototype through exponential moving average (EMA), so the final attentive prototype at each iteration $j$ is $f_{prototype} = \alpha f^{j - 1}_{prototype} + (1 - \alpha) f^j_{fusion}$, where $\alpha = 0.99$.

\noindent\textbf{Similarity-based Reweight.} \quad Finally, the classification loss is multiplied by these cosine similarity scores between each fused feature and the prototype representation. Since we use CenterPoint~\cite{yin2021center} as the original detector. It uses the class-balanced focal loss \cite{lin2017focal} as the classification loss with the calculation of classification scores prediction heatmap and ground truth heatmap. Therefore, the generation of weight $W \in [0, 1]^{w \times h \times c}$ is similar to the ground truth heatmap $Y \in [0, 1]^{w \times h \times c}$, where $w$, $h$, $c$ denotes the width, height and channels of the ground truth heatmap for classification. We use a modified Gaussian kernel function \cite{zhou2019objects} for each pseudo label $p$ as shown in the function below, where $W_{xyc}$ denotes the weight value in the location $(x, y)$ and channel $c$ for reweighting, $p_x$, $p_y$ denote the location of the pseudo label in heatmap, $s_{p}$ denotes the corresponding cosine similarity score and $\sigma$ is an object size-adaptive standard deviation.

$$
W_{xyc} = s_{p} \cdot \exp (-\frac{(x - p_x)^2 + (y - p_y)^2}{2\sigma^2})
$$

With this similarity-based reweighting, the losses will pay more attention to the corresponding regions that have been identified as correct through prototype matching and therefore the pseudo-label selection would bias to these consistent patterns, resulting in the final loss for target domain $L_{target} = W \cdot L^{cls}_{target} + L^{reg}_{target}$, which contains a class-balanced focal loss $L^{cls}_{target}$ for object classification, and a smooth-L1 loss $L^{reg}_{target}$ for bounding box regression. In this way, a soft-selection-based self-supervised framework is completed.

\subsection{Range Normalization}
\label{rn}

We also observe that mechanical LiDAR and solid-state LiDAR usually capture the scene with different perception ranges. Taking the mechanical LiDAR-based nuScenes dataset \cite{caesar2020nuscenes} and solid-state LiDAR-based PansaSet dataset \cite{xiao2021pandaset} for example, the former's perception range is [-50m, 50m], and the perspective view is a 360-degree ring area, while perception range of the latter is [0m, 100m], and the visual range is only a fan area of 60 degrees straight ahead. This difference in range also leads to a domain gap.

We propose range normalization (RN) as a data pre-processing to ease the difference in sensor range from different datasets. Specifically, we centralize all the non-centralized point cloud sample data to ensure that the origin of the point cloud coordinate system is in the center of the point cloud perception range. In other words, for PandaSet dataset whose original perception range is [0, 100m], we translate the overall point cloud data and corresponding annotated bounding box information, so that its perception range becomes [-50m, 50m] which is consistent with the data perception range of nuScenes dataset. It is a simple yet efficient normalization approach and considerable improvement can be observed in the 3D detection domain adaption task as shown in the ablation studies. 

\section{Experiment}
We first introduce all datasets and evaluation metrics used in the experiments and implementation details. After that, we explore cross-LiDAR domain shift scenarios and show compared 3D detection results to demonstrate the state-of-the-art performance of CL3D. Finally, we conduct extensive ablation studies to give a comprehensive assessment of submodules of CL3D.

\subsection{Experimental setup}
\label{exp_setup}

\noindent \textbf{Datasets} \quad We consider five widely-used large-scale autonomous driving datasets to simulate the various domain shifts, which are Waymo \cite{sun2020scalability}, nuScenes \cite{caesar2020nuscenes}, KITTI \cite{geiger2012we}, PandaSet \cite{xiao2021pandaset}, and PreSIL \cite{hurl2019precise}. Among them, Waymo is the largest dataset with more than 230K annotated 64-beam mechanical lidar frames collected across six US cities. nuScenes consists of 28130 training samples and 6019 validation samples collected by the 32-beam mechanical LiDAR and KITTI consists of 7,481 annotated lidar frames collected by the 64-beam mechanical LiDAR. PandaSet is the only dataset whose data is captured by solid-state LiDAR, including 5520 training samples and 2720 validation samples. Particularly, the synthetic dataset PreSIL contains 51075 synthetic LiDAR data generated from the Grand Theft Auto V (GTA V) game. Note that KITTI and PreSIL do not contain consecutive frames of data, so we will delete our TMA module for related experiments.

\noindent \textbf{Evaluation metric} \quad We adopt the nuScenes evaluation metric for evaluating our methods on the commonly used car category in most of our experiments. The Average Precision (AP) is used as the metric and a match is defined by thresholding the 2D center distance $d$ on the ground plane, then we average over matching thresholds of $d = \{0.5, 1, 2, 4\}$ meters and get the mean Average Precision (mAP). As for the synthetic-to-real domain adaptation from PreSIL to KITTI, we use the KITTI evaluation metric of intersection over union (IOU) of 0.7 in order to align other methods' performance. Refer to \cite{yang2021st3d}, we use \textbf{closed gap} $= \frac{AP_{model} - AP_{DT}}{AP_{Oracle} - AP_{DT}}$ to report how much the performance gap between Direct Transfer(DT) to Oracle is closed.

\noindent \textbf{3D detection network} \quad We use the CenterPoint detector \cite{yin2021center} as the base network, which is a one-stage anchor-free 3D object detector. A motion head is attached for extracting motion features by the supervision of object velocities (position offset between two adjacent frames) during the training on the source domain. The classification loss is reweighed during the computation in the domain adaptation process.

\noindent \textbf{Implementation details} \quad As for the implementation, we use the public pyTorch \cite{paszke2019pytorch} repository MMDetection3D \cite{mmdet3d2020} and we perform experiments with a 24GB GeForce RTX 3090 GPU. During both the pre-training and self-training processes, we adopt the widely adopted data augmentation, including random flipping, scaling, and rotation. The source data in pre-training process are trained for 20 epoch and target data in self-training process are trained for 1 epoch. Other settings are the same as official implementation of CenterPoint.

\subsection{Performance}
\label{exp_performance}

We mainly demonstrate the performance on cross-LiDAR 3D detection domain adaptation tasks. We compare with several approaches, including Direct Transfer (DT), Self-Training (ST), and current published SOTA ST3D~\cite{yang2021st3d} by running its released code on our experimental settings. As for another SOTA work SRDAN~\cite{zhang2021srdan}, we compare with it by aligning the performance on synthetic-to-real domain adaptation task according to its reported results for fair comparison. In particular, Direct Transfer(DT) indicates directly evaluating pre-trained model from the source dataset on the target dataset, and Self-Training(ST) indicates re-training the object detector supervised only by the pseudo-label generated by source-model. For adaptation experiments between mechanical LiDAR and solid-state LiDAR, we also apply Range Normalization (RN) to other SOTA methods. For other cross-domain settings without perception range gap, we will ignore the RN procedure.

\begin{table}[t]
\centering
\caption{Comparison on 3D car detection adaptation task between mechanical LiDAR dataset (nuScenes and Waymo) and solid-state LiDAR dataset (PandaSet). Oracle indicates the fully supervised model trained on the target dataset standing for upper bound of performance after adaptation.}
\label{tab_crossdataset}
\setlength{\tabcolsep}{2mm}{
\begin{tabular}{cc|c|c|c}
Source  & Target  &  Method   & mAP   & Closed Gap        \\ \hline \hline
nuScenes       & PandaSet       & DT  & 0.363   &   -    \\
               &                & ST & 0.379    &  2.60\%    \\
               &                & ST3D                 & 0.617    &  57.21\%    \\
               &                & CL3D        & \textbf{0.705} & \textbf{77.03\%} \\ \hline
               &                & Oracle                & 0.807    &  -    \\ \hline \hline
PandaSet       & nuScenes       & DT   & 0.089  &     -   \\
               &                & ST                & 0.244 &    27.63\%     \\
               &                & ST3D                & 0.349    &  46.35\%    \\
               &                & CL3D        & \textbf{0.467} & \textbf{68.98\%} \\ \hline
               &                & Oracle                & 0.650    &  -    \\ \hline \hline

Waymo          & PandaSet       & DT   &    0.272   &  -  \\
               &                & ST                & 0.383 &  20.75\%       \\
               &                & ST3D                 & 0.704   &  80.75\%    \\
               &                & CL3D        & \textbf{0.729} & \textbf{86.42\%}  \\ \hline 
               &                & Oracle                &  0.807   &  -   \\ \hline \hline
PandaSet       & Waymo          & DT   &   0.115   & -  \\
               &                & ST                & 0.312    &  37.10\%   \\
               &                & ST3D                & 0.423   &  58.00\%    \\
               &                & CL3D        & \textbf{0.492} & \textbf{71.00\%}  \\ \hline
               &                & Oracle                &  0.646  &  -   \\ \hline \hline

\end{tabular}
}
\vspace{-2ex}
\end{table}

\begin{table}[t]
\centering
\caption{Comparison on 3D car detection adaptation task between mechanical LiDARs with different beams, where nuScenes is 32-beam and KITTI is 64-beam.}
\label{tab_beams}
\setlength{\tabcolsep}{2mm}{
\begin{tabular}{cc|c|c|c}
Source  & Target  &  Method   & mAP   & Closed Gap        \\ \hline \hline
nuScenes       & KITTI          & DT   & 0.395      &  -  \\
               &                & ST                & 0.605     &  47.83\%  \\
               &                & ST3D                & 0.625    &  52.39\%  \\
               &                & CL3D        & \textbf{0.682} & \textbf{65.38\%}  \\ \hline
               &                & Oracle                & 0.834    &  -    \\ \hline \hline
KITTI          & nuScenes       & DT   & 0.116      &  -  \\
               &                & ST                & 0.179 &    11.80\%     \\
               &                & ST3D                 & 0.289    & 32.40\%     \\
               &                & CL3D        & \textbf{0.305} & \textbf{35.39\%}  \\ \hline 
               &                & Oracle                & 0.650    &  -    \\ \hline \hline

\end{tabular}
}
\vspace{-1ex}
\end{table}

\begin{table}[h]
\centering
\caption{Comparison of different methods under the synthetic-to-real scenario from PreSIL to KITTI dataset using the metric mAP adopted by KITTI. ‘-’ indicates the results are not available in their works.}
\label{tab_comparison}
\setlength{\tabcolsep}{2.3mm}{
\begin{tabular}{c|ccc}
\multirow{2}{*}{PreSIL -\textgreater KITTI} & \multicolumn{3}{c}{mAP(Car)}                  \\ \cline{2-4} 
                                            & Easy          & Mod.          & Hard          \\ \hline \hline
DABEV                                     & -             & 17.1          & -    \\
CDN                                         & -             & 19.0          & -             \\
SWDA-3D                                     & 22.6          & 18.7          & 16.3          \\
SRDAN                                      & 25.9          & 22.1          & 18.7          \\ \hline \hline
CL3D                                  & \textbf{28.0} & \textbf{25.3} & \textbf{23.4} \\ \hline \hline
\end{tabular}
}
\vspace{-1ex}
\end{table}

\begin{table}[ht]
\centering
\caption{Ablation study for SGA and TMA of our method.}
\label{tab_modular}
\setlength{\tabcolsep}{2.5mm}{
\begin{tabular}{c|c|c|c|c}
Method                                 & \makecell[c]{w/o \\ TMA \& SGA} & \makecell[c]{w/o \\ TMA} & \makecell[c]{w/o \\ SGA} & CL3D \\ \hline \hline
mAP                  & 0.641 & 0.686 & 0.662 &   \textbf{0.705}        \\
\end{tabular}
}
\end{table}

\begin{table}[t]
\caption{Comparison of different solutions for range gap. }
\centering
\label{tab_range_norm}
\begin{tabular}{c|c|c|c|c}
Method                      & ST & ST+RSym & ST+RSp & ST+ RN  \\ \hline \hline
mAP                          & 0.379  & 0.361       & 0.343  & 0.641
\end{tabular}
\vspace{-2ex}
\end{table}

Table.~\ref{tab_crossdataset} shows the adaptation results between solid-state LiDAR dataset PandaSet and mechanical LiDAR datasets, including nuScenes and Waymo, and our method attains an obvious improvement over other methods. ST is superior to DT because of the retraining under the guidance of pseudo labels. Our method outperforms ST due to the domain alignment process with the assistance of domain-invariant geometry and motion features. ST3D utilizes threshold strategy to select high-quality pseudo-labels, which may lead to involving incorrect labels that have high confidence and discarding correct labels with low confidence in the self-training procedure. Our method is superior to ST3D, which is mainly due to the soft selection mechanism by reweighting the pseudo labels via prototypes, which avoids the misjudgement of pseudo labels by hard constraints and further benefits the learning of the network. Furthermore, to demonstrate the effectiveness of our method on the domain shift caused by different beams of mechanical LiDARs, we conduct the experiment between KITTI and nuScenes in Table  \ref{tab_beams} and also get the state-of-the-art performance.

\begin{figure}
    \centering
    \includegraphics[scale=0.26]{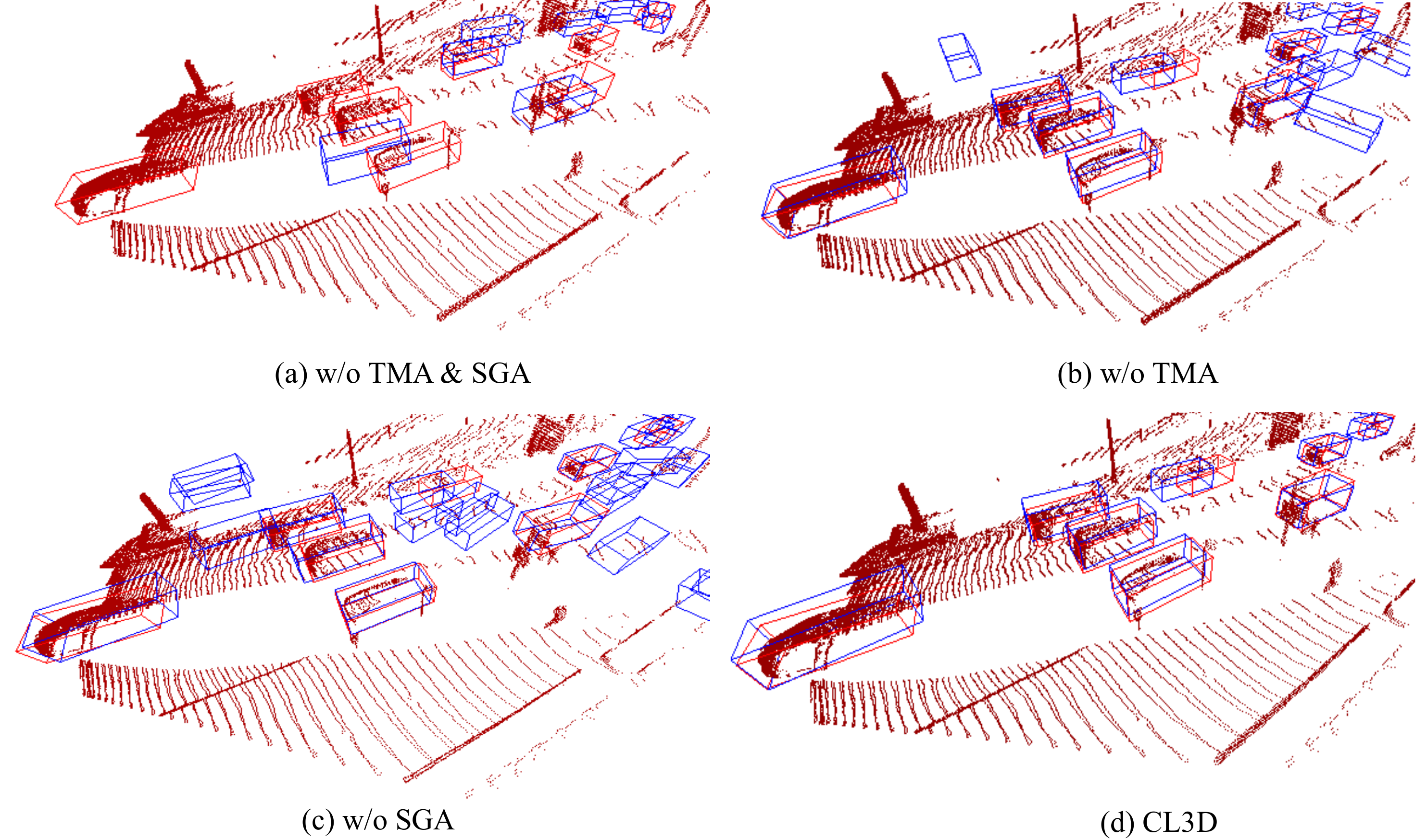}
    \vspace{-1ex}
    \caption{Visualization of 3D detection for the domain adaptation from nuScenes to PandaSet, where red boxes are ground truth and blue boxes are predictions.}
    \label{fig_res}

\end{figure}

\begin{table}[t]
\centering
\caption{Ablation study for different backbones of detector.}
\label{tab_backbone}
\setlength{\tabcolsep}{0.8mm}{
\begin{tabular}{c|c|c|c}
\diagbox[]{\small Method}{\small Backbone} & voxel-based & pillar-based & point-based \\ \hline
DT     & 0.363       & 0.247        & 0.624       \\
ST     & 0.379       & 0.303        & 0.632       \\ \hline \hline
CL3D   & 0.686       & 0.445        & 0.691      
\end{tabular}
}
\end{table}

\begin{table}[]
\caption{Results of CL3D on more challenging categories, where tr. represents the truck category and pe. represents the pedestrian category.}
\centering
\label{tab_more_cat}
\begin{tabular}{c|c|cc}
Source $\rightarrow$ Target & Method   & mAP(tr.) & mAP(pe.) \\ \hline
nuScenes $\rightarrow$ PandaSet       & DT   & 0.000          & 0.030            \\
                  & ST        & 0.096      & 0.086           \\
                  & CL3D        & 0.131      & 0.124           \\ \hline
                  & Oracle            & 0.284      & 0.295           \\ \hline \hline
PandaSet $\rightarrow$ nuScenes       & DT   & 0.000          & 0.003           \\
                & ST        &    0.072       &       0.153         \\
                & CL3D        &       0.113     &       0.294          \\ \hline
                & Oracle            & 0.275      & 0.524          
\end{tabular}
\vspace{-2ex}
\end{table}

Beyond cross-LiDAR domain adaptation problems, synthetic-to-real domain adaptation is also significant due to the difficulty in collecting and annotating large-scale data in real-world scenarios. Therefore, we further validate our method under the synthetic-to-real setting using the synthetic PreSIL dataset and real KITTI dataset. We use the IOU metric of KITTI in order to align other methods' results~\cite{zhang2021srdan,saito2019strong,su2020adapting,saleh2019domain}. The results are shown in Tabel \ref{tab_comparison}. Our method outperforms all methods by convincing margins, indicating that our method is also suitable for bridging the domain gap between the synthetic and real domain. That is because, no matter the cross-LiDAR domains or the synthetic-to-real domains, domain-invariant features are all about the geometric characteristics and motion patterns and our method exploits the invariant information to solve the domain gap essentially.

\subsection{Ablation studies}
\label{exp_ablation}

To evaluate the effectiveness of submodules of our method, we conduct ablation studies and analyze their contributions to the unsupervised domain adaptation task from nuScenes to PandaSet. We also show the performance of using different detector backbones and on other categories of objects to further illustrate our method's generalization capability.

\noindent \textbf{Effectiveness of SGA and TMA} \quad We first study the effects of spatial geometry alignment (SGA) and temporal motion alignment (TMA). Table \ref{tab_modular} and Figure.~\ref{fig_res} show quantitative and qualitative results, respectively. SGA extracts the domain-invariant geometry-aware features and TMA learns the motion-aware features for the same kind of objects in different domains, which are all used during the self-training process to generate prototype representation for specific class to select high-confidence pseudo labels with soft constraints. It is obvious that both of them play critical roles.

\noindent \textbf{Effectiveness of Range Normalization} \quad To solve the huge range gap between mechanical LiDAR and solid-state LiDAR, we design RN in data pre-processing for the range alignment. It's simple but effective. Actually, we have tested several solutions to reduce the range gap, as Table.~\ref{tab_range_norm} shows. Range Symmetrization (RSym) copies the fan area of solid-state LiDARs to complete the symmetrical area as mechanical LiDARs. Range Split (RSp) splits the circular area of mechanical LiDARs into fans to align solid-state LiDARs. These intuitive methods do not work and even have negative effects, while RN produces significant performance gains by regularizing all point cloud to a general range, which eases the learning difficulty and has good generalization capability for current LiDAR categories.

\noindent \textbf{Different detection backbones} \quad Additionally, we show the performance of using different backbones in the CenterPoint detector, namely voxel-based, pillar-based and point-based backbones, which are three main types of backbones in 3D perception area, to verify the generalization capability of CL3D. Voxel-based backbone utilizes the structured voxel representation to quantize the LiDAR data while pillar-based backbone utilizes the pillar representation for efficient point process. Point-based backbone extract features from raw point cloud data directly. Results in Table \ref{tab_backbone} demonstrate that our method can boost the domain adaptation performance based on different detector backbones. Among them, the voxel-based backbone is a good choice with high-efficient processing for large-scale point cloud and precise detection performance.

\noindent \textbf{More challenging categories} \quad Except for the normal car category concerned in most 3D detection domain adaptation methods, we also conduct experiments on other two important types of traffic agents, including trucks and pedestrians. Table.~\ref{tab_more_cat} shows experimental results. Direct Transfer can hardly obtain predictions. Our method gets improvement by a large margin, demonstrating that our method is solid for different types of detection objects. Specifically, due to the limited training samples on these challenging categories, even the result of oracle is not good.

\section{Conclusions}
We propose an unsupervised domain adaptation method to bridge the domain gap for LiDAR-based 3D detection caused by the differences in perception range, point cloud density, and point arrangement. In particular, we design Spatial Geometry Alignment to extract similar 3D shape geometric features and Temporal Motion Alignment to extract similar motion patterns of the same category from distinct instance-level distributions to align two domains. Extensive experiments and comprehensive ablation study demonstrate the effectiveness of our approach for cross-LiDAR 3D object detection. Although prototype representation solves the false classification, the deviation caused by scale and location error still exists, which we aim to solve in the future.

\section{Acknowledgements}

This work was supported by NSFC (No.62206173), Shanghai Sailing Program (No.22YF1428700), and Shanghai Frontiers Science Center of Human-centered Artificial Intelligence (ShangHAI).

\clearpage
\bibliography{aaai23}

\end{document}